\documentclass{article}
\usepackage{graphicx}

\PassOptionsToPackage{numbers, compress}{natbib}
\usepackage[final]{neurips_2022_ml4ad}

\usepackage[utf8]{inputenc} 
\usepackage[T1]{fontenc}    
\usepackage{hyperref}       
\usepackage{url}            
\usepackage{booktabs}       
\usepackage{amsfonts}       
\usepackage{nicefrac}       
\usepackage{microtype}      
\usepackage{xcolor}         

\usepackage{bbm}

\title{One-Shot Learning of Visual Path Navigation for Autonomous Vehicles}

\author{%
Zhongying CuiZhu$^{1}$\thanks{These authors contributed equally to this work.} \quad Francois Charette$^{1*}$ \quad Amin Ghafourian$^{2*}$ \quad Debo Shi$^2$ \quad Matthew Cui$^3$\thanks{The authors contributed to this work while at Ford Greenfield Labs.} \\
\textbf{Anjali Krishnamachar}$^{4\dagger}$ \quad \textbf{Iman Soltani}$^2$\thanks{Corresponding author.}\\
$^1$Ford Greenfield Labs \quad $^2$University of California, Davis \quad $^3$Google \quad$^4$ MIT Sloan\\
\texttt{\{zzhou54, fcharett\}@ford.com}\\
\texttt{\{aghafourian,deshi,isoltani\}@ucdavis.edu}\\
\texttt{mattcui@google.com} \quad
\texttt{anjalik@mit.edu}
}
\begin{document}

\maketitle

\begin{abstract}
  Autonomous driving presents many challenges due to the large number of scenarios the autonomous vehicle (AV) may encounter.  End-to-end deep learning models are comparatively simplistic models that can handle a broad set of scenarios.  However, end-to-end models require large amounts of diverse data to perform well.  This paper presents a novel deep neural network that performs image-to-steering path navigation that helps with the data problem by adding one-shot learning to the system. Presented with a previously unseen path, the vehicle can drive the path autonomously after being shown the path once and without model retraining. In fact, the full path is not needed and images of the road junctions is sufficient. In-vehicle testing and offline testing are used to verify the performance of the proposed navigation and to compare different candidate architectures.
\end{abstract}

\section{Introduction}

Recent advances in machine learning result in high performance on tasks like classification, but at the cost of requiring large amounts of training data \cite{imagenet} \cite{vit}. The data problem is especially important in the autonomous driving domain because the outputs of these models are used for safety critical applications and therefore must have high accuracy and cover a wide range of cases. This means that autonomous driving models may require thousands or even millions of annotated data examples, which can be very difficult to acquire and time consuming to annotate and pre-process \cite{bojarski2016end}. Other problems include the difficulty of collecting abundant data for edge cases that occur infrequently, and then balancing the dataset so that the model performs well for both edge cases and normal scenarios. Futhermore, once a model is trained, it is difficult for the model to adapt to new scenarios without offline retraining. We attempt to address these data problems with a novel approach that combines an end-to-end autonomous driving system with the adaptability of one-shot learning. 

End-to-end models are a promising approach to autonomous driving and offer a comparatively simplistic design to conventional systems \cite{bojarski2016end}. With the success of NVIDIA’s PilotNet in 2016, a competitive image-based end-to-end steering system, interest has increased further for this approach \cite{bojarski2016end}, \cite{MTLsteer}, \cite{Solomon2019HierarchicalMD}. However performance of end-to-end systems, like other deep learning models, is predicated on large amounts of diverse data.  Another drawback is that end-to-end networks are a black box and therefore extremely difficult to diagnose what prompts inference failures. Although explainability can be important for safety critical applications, this work values the simplistic design and portability of the end-to-end steering system to an autonomous vehicle. Still, the hierarchical multi-task network \cite{Solomon2019HierarchicalMD} that is used as the basis of our autonomous driving network is modular by design and allows for some interpretability.

Few-shot learning is a facet of meta-learning, or ``learning how to learn''.  Under this framework, models can learn new tasks or adapt to new environments rapidly with a few training examples \cite{Lakemeta}.  At one extreme, one-shot learning models can learn from just one example.  Few-shot learning has been applied to several tasks ranging from classification \cite{memorynetworks}, \cite{siamese2015}, \cite{matching2016} to reinforcement learning \cite{LearningRL}, \cite{finnmaml}.  Few-shot learning has been implemented as part of robot navigation to change color features of a target \cite{fewshotR}, and for road object detection \cite{roadobject}. However, to the best of the authors' knowledge, no previous work uses few-shot learning for image recognition as part of an autonomous driving system. 

In this paper, we develop a novel end-to-end deep learning architecture with one-shot learning to accomplish visual path navigation for autonomous driving. The main contributions of this work are:
\begin{enumerate}
\item We develop an end-to-end autonomous driving architecture that can learn navigation tasks in one-shot. Various architecture designs were considered and hyperparameter tuned to result in our proposed model. After a user drives the vehicle through a new path once, the vehicle can replicate the path autonomously. In fact, with our setup, a full drive through the new path is not needed and reference images of the junction turns are enough. Learning new paths do not affect the system's performance on previously trained paths.

\item We relax the need for large amounts of training data compared to classifier-based baselines and the hierarchical multi-task network, which our network builds upon \cite{Solomon2019HierarchicalMD}.

\item We implement and test the proposed autonomous driving system and one-shot learning in real-time with a mini-Autonomous Vehicle (AV) in both indoor and outdoor settings. The mini-AV can also collect and automatically label training and testing data.

\item We build a custom metric to compare the performance between candidate model. The proposed metric combines the temporal and discrete aspects of the task and generates a combined score. It improves the descriptiveness of metrics such as simple accuracy statistics or confusion matrices for this task.
\end{enumerate}

 As this work is a proof of concept, we use a simplistic end-to-end driving model and limited environments for training and testing. But there are still many applications of our system.  For example, because our model only requires images of junctions where the vehicle turns, it can navigate new paths by downloading images of intersections from online maps. 
 Furthermore, with a different driving network, this approach could be generalizable to more complicated paths.
 

\section{Related Work}

\subsection{One-Shot Learning}

There are various approaches to one-shot learning. Many approaches use memory and metric-based techniques \cite{memorynetworks}, \cite{siamese2015}, \cite{matching2016} \cite{prototypical} developed for image classification tasks or language applications.  In memory-augmented neural networks \cite{memorynetworks}, memories are stored externally and compared against the test image through cosine similarity.  In Siamese networks \cite{siamese2015}, there are two CNN branches that share weights.  One branch processes the test image and the other processes the reference image, and the resulting vectors are compared using a component-wise weighted L1 metric for image recognition.  The relational network \cite{relational2018} uses the same basic Siamese architecture, but replaces the L1 metric with a simple concatenation of the two feature vectors followed by a CNN.  In another method built on Siamese networks, matching networks \cite{matching2016} use separately learned networks to embed the test and reference images.  Each of these embedding networks is based on an LSTM that generates the output embedding in the context of the other classes.  
Our proposed model draws inspiration from these models and uses the Siamese network to produce feature vectors, of which reference image features are stored in memory. A subsequent module compares two image sequences instead of two images.

\subsection{Hierarchical Multi-Task DNN \cite{Solomon2019HierarchicalMD}}
The foundations for this work are based on development of the novel hierarchical multi-task deep neural network (DNN) in  \cite{Solomon2019HierarchicalMD}. The multi-task model is based in part on the concept of Multi-Task Learning in \cite{MTLsteer}, \cite{MTLDNN}, \cite{MTLAV}, \cite{MTLE2E} and proposes a modular end-to-end architecture that contains two levels of hierarchy: a high-level manager classifier network and a number of worker task models. This is a generic architecture that can be used on any application that can be partitioned into sub-tasks. The manager classifier selects which task to perform and the output from that task becomes the final model output. For an autonomous driving application, each task represents a driving action such as "turn left" or "turn right", and outputs steering commands that complete the appropriate action. The task models are implemented as regression networks. 
This overall architecture is the basis for the steering control system of the small autonomous driving platform used in this research effort.

\subsection{Route Following}
There are several approaches to route following for robots.  One approach learns the map from a training run and the map is used for localization in subsequent runs \cite{teachnrepeat}.  Another approach uses odometry data with periodic orientation correction using visual input \cite{bioteach}.  Still another considers using teaching runs from another platform and solving the viewpoint mismatch problem \cite{voila}.  Our problem differs from the aforementioned because we solve an autonomous driving problem, where vehicles drive on well-defined roads, as opposed to open-world path navigation for robots.  As such, it is difficult to directly compare the path following capability between the two.  Also, our approach uses purely visual inputs and need only images of road junctions instead of a full training drive.

\section{Formulation of One-shot Problem}
\subsection{Mini-Autonomous Vehicle Platform}
One contribution of this paper is to set up a real-time mini-AV platform, on which data is collected and the driving algorithm is implemented and tested \cite{Solomon2019HierarchicalMD}. This hardware platform can simultaneously record the steering and throttle commands from the remote controller (RC), and the stereo images from the camera at a rate of approximately 30 frames-per-second, which are automatically synchronized generating automatically annotated data.

\begin{figure}[thpb]
      \begin{center}
      \includegraphics[width=.8\columnwidth]{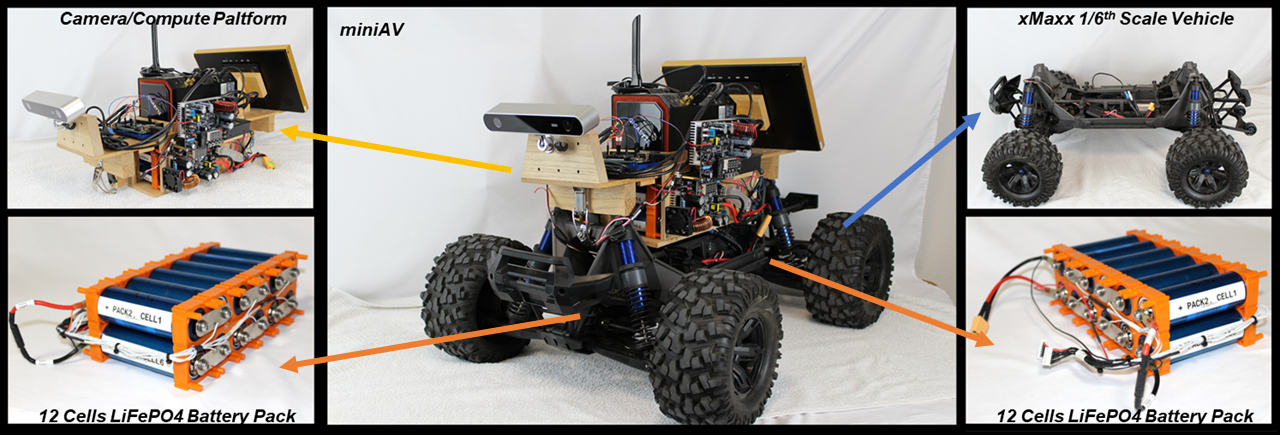}
      \caption{Miniature Autonomous Vehicle modified from a Remote Control car}
      \label{fig:miniAV}
      \end{center}
    \end{figure}

A 1/6 scale Traxxas xMaxx vehicle shown in Figure \ref{fig:miniAV} was used.  The vehicle was modified to operate autonomously with a stereo camera, an Arduino micro-controller, an Intel processor with GTX-970 graphics card with 4 GB of memory, and a Vedder electronic speed controller. A 12 cells battery pack was mounted to provide power, and an external monitor was installed.

The vehicle has three driving modes: manual, recording and autonomous.  In manual mode, the vehicle behaves like a standard RC car and the user controls the throttle speed and steering angle.  In recording mode, the vehicle is still user-controlled but the stereo image data, and the corresponding steering angle and throttle speed are recorded frame-by-frame \cite{hdf5}.  In autonomous mode, the vehicle is fully autonomous and the Intel processor controls the throttle speed and steering angle based on real-time camera input. The vehicle runs at 15 frames-per-second in our proposed navigation system.  

\subsection{One-shot Learning Formulation}
The classical formulation for the few-shot learning classification problem is as follows.  $\mathcal{C}$ is the set of classes for all $y_i$ such that $y_i \in \mathcal{C}$. For a set of novel classes $\mathcal{C}_{novel}$ to be classified, a set of base classes $\mathcal{C}_{base}$ is used for training, where $\mathcal{C}_{novel} \cap \mathcal{C}_{base}=\emptyset$ and $\mathcal{C}_{novel} \cup \mathcal{C}_{base}=\mathcal{C}$.  The training dataset is $\mathcal{D}=\{(\mathbf{x}_i,y_i)\}, \mathbf{x}_i\in \mathbb{R}^d, y_i\in \mathcal{C}_{base} $.  The goal is to train a model with training data from the base classes so that the model can generalize well on tasks sampled from the novel classes.  In the evaluation phase of $K$-shot learning, a series of tasks $\{\mathcal{T}\}$ are evaluated. 
Each task is split into a support set $\mathcal{S} = \{(\mathbf{x}_i, y_i)\}^{K\times N}_{i=1}, y_i\in \mathcal{C^T}$ and a query set $\mathcal{Q} = {(\mathbf{x}_i, y_i)}^{K\times Q}_{i=1}, y_i \in \mathcal{C^T}$, where $N$ is a subset of $\mathcal{C}_{novel}$.   
$K=1$ in the case of one-shot learning and $N=Q$ for our proposed one-shot setup. An evaluation task for the proposed setup is a novel path driven by the mini-AV. 

\subsection{Navigation as a One-shot Problem}
\begin{figure}[thpb]
      \begin{center}
      
      \includegraphics[width=.5\columnwidth]{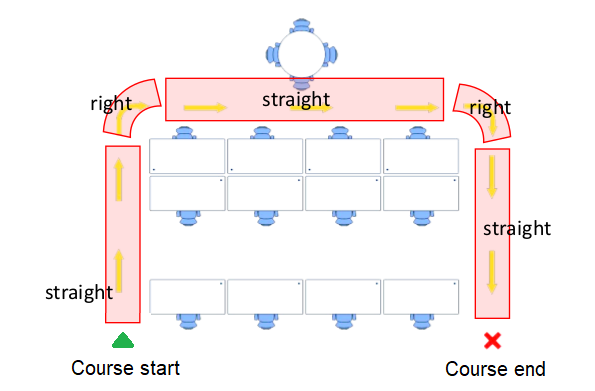}
      \caption{Example of a course split into separate tasks/sections}
      \label{fig:deskPath}
      \end{center}
   \end{figure}

The hierarchical multi-task DNN \cite{Solomon2019HierarchicalMD} is the basis model for our proposed autonomous driving architecture. Following their formulation, driving is partitioned into "left", "right" and "straight" tasks or sections. An example of this partition is shown in Figure \ref{fig:deskPath}.  Each task is implemented with a regression network that outputs a steering command based on input images. The manager classifier network (MCN) that selects which task to perform is implemented as a Siamese network to allow one-shot learning.

The MCN is a multi-input binary classifier where its input is a stream of images, and its binary output signals when a task is complete. The next task is loaded from a memory queue and the process repeats as shown in Figure \ref{fig:MCN} and discussed in detail in Section \ref{model_arch}. Each task is stored as a direction and a sequence of 10 images taken from the end of the task, which is compared against the input image stream in a Siamese network. 
The order of the memory queue is the order of tasks that the mini-AV encounters when driving from course start to end as shown in Figure \ref{fig:deskPath}. The initial state is assumed to be the "straight" task. The mini-AV stops when there are no more tasks in the memory queue.

To train the network, multiple manual runs of the same course are recorded, one of which provides the reference images while the rest serve as the image input stream. Different combinations of the reference and input images are generated to produce positive and negative examples for the MCN to learn how to recognize the end of a section.  

Once the network is trained, the weights in the model are frozen for one-shot learning. One-shot learning for path recognition is tested as follows on the mini-AV testing platform:
\begin{enumerate}
  \item Learning: The mini-AV learns a particular path by being manually driven in recording mode once. The path is automatically annotated into different sections, with the correct task label and the last 10 images of the associated section saved in order in a memory queue.
  \item Inference: The mini-AV is placed in autonomous mode with the memory queue loaded. The mini-AV drives autonomously to repeat the learned path. 
\end{enumerate}

\section{Model and Data}
\subsection{Model Architecture} \label{model_arch}
\begin{figure*}[thpb]
      \begin{center}
      \includegraphics[width=.95\textwidth, height=6cm]{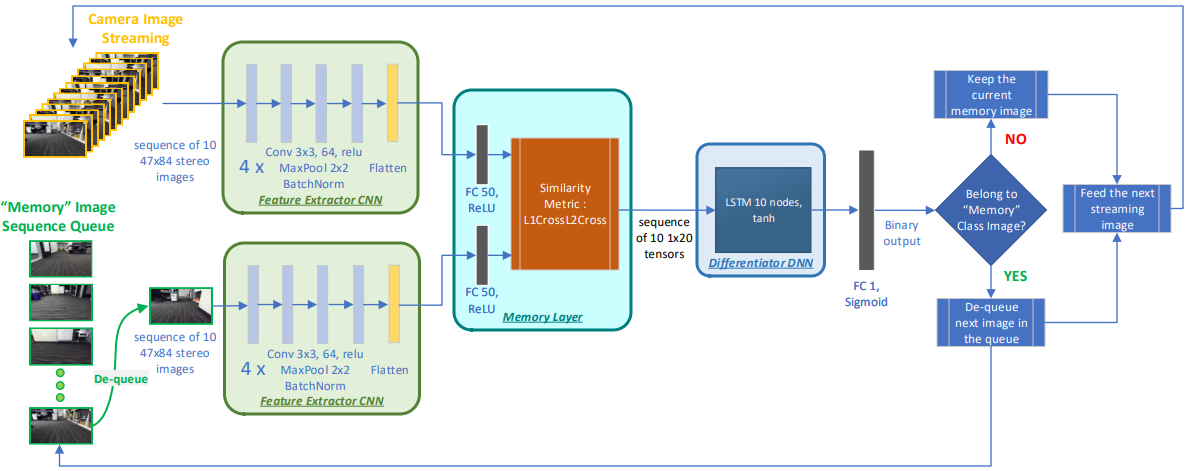}
      \caption{MCN and its three components: feature extractor, memory layer, and differentiator}
      \label{fig:MCN}
      \end{center}
    \end{figure*}

The Manager Classifier Network (MCN) is a multi-input binary classifier. The MCN takes as input a stream of images (such as a real-time camera stream) and a reference image sequence and determines if the two match. For this driving application, the reference image sequence and input image sequence are a sequence of 10 images. The MCN is comprised of the feature extractor followed by the memory layer and finally the differentiator.  The overall structure is shown in Figure \ref{fig:MCN}.

Two identical feature extractor networks are arranged in a Siamese fashion where a test image is input into one branch and a reference image is input into the other branch. The corresponding weights in the two feature extractors are identical. The feature extractor is composed of a convolutional neural network (CNN) followed by a dense layer.  The CNN is composed of four consecutive modules, each of which is a 3x3 convolution with 64 ﬁlters followed by a ReLU activation, 2x2 max-pooling, and batch normalization. This Conv-4 configuration is a competitive and simple backbone often used in few-shot learning \cite{siamese2015} \cite{constellationnet} \cite{du2022hierarchical}.  A dense layer compresses the CNN output into a 50-dimensional embedding of the image.  For a sequence of input images, each image is run through the feature extractor network and the overall output is a tensor of the concatenated feature vectors. 

The memory layer compares the two feature vector sequences using a similarity metric. The similarity can be measured with standard techniques such as cosine similarity \cite{matching2016}, component-wise L1 \cite{siamese2015}, and L2 distances or with more sophisticated metrics \cite{wasserstein} \cite{distance}, and is a design choice. The various similarity metrics are then concatenated into a tensor and input to the differentiator.

The differentiator takes the inputs from the memory layer and produces a binary output signaling whether the reference and test images match. The differentiator can be implemented as several fully connected layers, RNN, LSTM, or transformers \cite{rnn0} \cite{lstm0} \cite{transformer}. Through experiments and analysis, an LSTM with 10 hidden nodes performs best. The loss function is a standard binary cross-entropy loss.

\subsection{Data Collection}
One-shot learning requires the training data to consist of diverse tasks that relate to the target application. Training data for path navigation is collected as the user manually drives the vehicle through a course in recording mode. For convenience, we trained and tested in an indoor office setting consisting of two buildings, building A and building B. We recorded training data from building B and tested the vehicle in building A. Both buildings contain clusters of desks, chairs and cabinets, but are set up differently to present a generalization challenge to our network.  

For training, we recorded 13 courses in total with each course consisting of 8-15 turning sections.  We recorded 4 runs per course, 3 of which are used in training and one for testing. Mirrored versions of each course augment the data for a total of 26 distinct courses.  For one-shot testing, we recorded 2 courses with 2 runs per course, so that we can test in real-time as well as in simulation.

A small outdoor dataset was also collected to demonstrate the model in another environment. Due to limitations, we recorded 2 training courses with 3 runs per course and 1 test course with 2 runs per course. The rest of the data processing and model hyperparameter selection was tuned for the bigger indoor dataset then applied to the outdoor dataset.  Further details can be found in Section \ref{outdoor}.

\subsection{Data Pre-processing}
The raw stereo images are resized from $672\times 376$ to $84\times 47$ with each image having 3 color channels for a total of 6 color channels per stereo image. The raw steering values are linearly mapped to normalized steering values between $-100$, a full left turn, to $100$, a full right turn.  Each image frame is labeled as belonging to a ‘left’, ‘right’ or ‘straight’ section based on the normalized steering values. Further information about this procedure is available in the appendix.

\subsection{Data Pairing for Training}
Each training example is composed of a pair of reference and test sequences, a consecutive sequence of ten frames each.  
From 13 indoor training courses, a total of \textasciitilde 300,000 training examples are created, of which 15\% is randomly set aside as validation data. Similarly, 2 outdoor courses result in \textasciitilde 45,000 training examples.  Indoor and outdoor models are trained separately. 
A training batch size of 8 is used and each batch contains randomly shuffled positive and negative examples from the same section of the same course. 

The positive and negative examples are constructed as follows. 
 Positive examples are formed by pairs of reference and test sequences from the same section end. Negative examples are pairs where the reference sequence is from the section end but the test sequence is not. The section end is the last 15 frames of a section.  A training buffer region of 10 frames before the section end is set aside to allow for slight differences during the manual drives. The number of negative examples is chosen to be double the number of positive examples because there is more diversity in the negative examples.

\section{Results}
After optimizing over hyperparameters and several architectures, we verified the proposed MCN model performance through both one-shot simulation testing as well as real world testing. The models were trained on one Nvidia Quadro RTX 6000 GPU for 10 epochs which took about 12 hours.

\subsection{Performance Metric}
\label{subs_perf}
We developed a custom performance metric to evaluate the one-shot learning model.  In our navigation approach, a false positive that triggers a task change and false negative that misses the task change will both cause failures.  Conversely, a section passes only if there are no early triggers, and there is a trigger at the section end. The prediction can achieve a 99\% accuracy and still have false triggers.  Thus, the raw accuracy is not informative enough and a new performance metric was created. This new metric evaluates the prediction section by section, and combines them into one numerical score.  

The section body accuracy of section $j$ in Equation \ref{eq:1} is based on the true negative rate or specificity, and is weighted so that an error early in the section body has more severe consequences.  $startframe_j$ is the first frame of the section body for section $j$ and $n_j$ is the last frame of the section body. $\{section end_j\}$ is the set of frames in the $j$-th section end. $p_i$ is the predicted output of the $i$th frame sequence.

\begin{equation} \label{eq:1}
        \textrm{section body accuracy}_j = \frac{\sum_{i = start frame_j}^{n_{j}}{(n_{j}-i)\cdot \mathbbm{1}[p_i < 0.5]}}{\sum_{i = start frame_j}^{n_{j}}{(n_{j}-i)}}
\end{equation}

The $j$th section end accuracy in Equation \ref{eq:2} is also the true positive rate or sensitivity of section $j$.
\begin{equation} \label{eq:2}
        \textrm{section end accuracy}_j = \frac{\sum_{i \in \{section end_j\}} \mathbbm{1}[p_i> 0.5]}{\textrm{ number of frames in section end}_j}\label{endAccuracy}
\end{equation}

The section body accuracy and the section end accuracy are each calculated per section within the test run, and the two accuracies are combined using a geometric mean into a per-section percentage accuracy. Next, a geometric mean is taken across all sections’ percentage accuracies. The overall score is penalized for error by halving the score for each section with an error (in section body or section end or both). That way, it is easy to spot if a model fails in at least one of the sections of the test run, since the performance score will be $< 0.5$.  It is possible for all sections to pass and still have a performance score $< 0.5$, however this means the pass is low quality, and slight fluctuations may cause the model to fail.  Thus, we aim to achieve a high performance score of above $0.5$.  

\subsection{Data Augmentation}
One-shot learning applications are trained on a small amount of data, so augmenting the available data can help the model generalize better. In this paper, various augmentation methods such as mirroring, cropping, and rotating training videos were tested.  Each augmented course is considered a separate course from the original counterpart.  In addition, doubling the amount of near-negative training examples was also tested, where a near-negative example is defined as a training example where the reference images are in the section end and the test images are within 15 frames of the section end.  Mirroring and adding near-negative examples were found to help with the model performance whereas rotation caused the model training to diverge, the reason for which may be that the rotated images were too different from the original.

\subsection{Model Studies}
We conducted studies by implementing the feature extractor with 2D and 3D CNNs. For the 2D CNN models, we investigated the effects of number of layers, filter size, and number of filters. We confirmed the commonly used Conv-4 architecture with 4 layers, 3$\times$3 filter size and 64 filters performs the best with our one-shot task with real-time considerations \cite{siamese2015}.

We also tried an alternate 3D convolutional architecture. 3D convolutions incorporate the temporal aspect and performs well in video applications \cite{conv3D}.  We modeled our architecture after the original 3D convolutions paper and also took cues of parameter selection from their experiments. We implemented several metrics for the memory module.  The differentiator was modified to densely connected layers. The models trained but did not generalize well to the one-shot tests.  This may indicate that we need additional data for the larger 3D CNNs, and that perhaps compressing images in the time dimension adds an additional challenge when trying to compare two short sequences of images.

The memory layer compares two feature vector sequences using a similarity metric.  We found that a feature vector length of 50 gives the best performing model.  Out of the various metrics we tried, a concatenation of L1 and L2 distances applied to each pair of feature vectors, one from reference and one from input (similar to a covariance matrix), worked best. L2 has worked well in other few-shot models \cite{prototypical} so an extended version of L2 working well with our sequential data is reasonable. Probability based distances may further improve our model and remain as future work \cite{wasserstein} \cite{distance}.

The differentiator is the last component of our network and we found LSTM to work best among dense layers, RNN, LSTM, and transformers.  For the transformer network, instead of using a positional encoding, we used a time convolution at the beginning to encode temporal relationships \cite{transformer}. We varied the embedding activation, time convolutional kernel length, and number of heads. The best tuned transformer network results were comparable to the LSTM network results but was not consistent across the two one-shot tests. Still, the CNN+transformer architecture shows promise; perhaps a full transformer model including vision transformers could improve results in the future. 

A naive classifier baseline is also evaluated.  The Conv-4 based classifier learns each section end as a separate class. To learn a new course at test time, the last layer of the classifier is replaced with cosine similarity to compare with the reference section end images. The resulting scores are lower than the one-shot learning based models and demonstrate the effectiveness of one-shot learning when sparse data is available.  Tested architectures and their performance in simulated one-shot tests are summarized in Table \ref{table:1shot}.
\begin{table}[h]
\caption{Simulated one-shot testing results on 2 courses using our proposed metric from Section \ref{subs_perf}.}
\label{table:1shot}
\centering
\begin{tabular}{c|c|c}
    Methods & Test course 1 score & Test course 2 score\\
    \hline
    Naive classifier & 0.016 & - \\
    Transformer & 0.28 & 0.94  \\
   
    3D conv & 0.12 & 0.15 \\
    
    CNN+LSTM & \textbf{0.83} & \textbf{0.96}
\end{tabular}
\end{table}

\begin{figure}[h]
		\centering
		\includegraphics[width=.8\columnwidth]{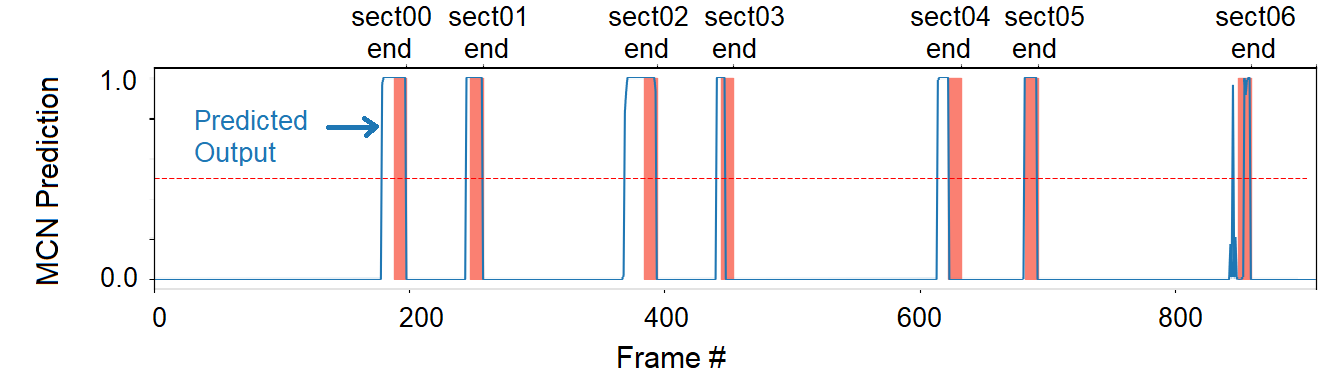}
    		\caption{ CNN-LSTM model one-shot simulation test on course 1. The blue prediction line appropriately triggers at each section end (red bars) within an allowable lead-up of 15 frames.}
		\label{fig:f_bestrun}
\end{figure}

The simulation results of the CNN+LSTM model for test course 1 are shown in Figure \ref{fig:f_bestrun}. The x-axis is the frame number of the input image stream, and the y-axis is the prediction results. The blue line predicts whether the input image sequence matches the reference. There is a test buffer of 15 frames before the section end. It can be seen that each section end is appropriately triggered.

We also verify that the simulation results translate to real-time results by letting the mini-AV autonomously drive through the two one-shot test courses.  

\subsection{Data Results}


The Hierarchical network in \cite{Solomon2019HierarchicalMD} required more runs of a course compared to our one-shot version to learn path navigation as shown in Table \ref{table:numruns}.  This, along with the poor performance of the naive classifier baseline supports the hypothesis that one-shot learning models help in training data reduction.

\begin{table}[h]
\caption{Number of runs required to learn path navigation.}
\label{table:numruns}
\centering
\begin{tabular}{c|c|c}
      & Hierarchical DNN & One-shot enabled\\
    \hline
    \# of runs/course & 4-10 & 3
\end{tabular}
\end{table}

Our model has the additional capability of learning new courses with one example run and without further training. This framework does not ``forget'' previous training courses as demonstrated by testing the model with a held out run of select training courses shown in Table \ref{table:traintest}.
\begin{table}[h]
\caption{Testing with held out runs of select training courses.}
\label{table:traintest}
\centering
\begin{tabular}{c|c|c|c}
     Method & Train Course 1 score & Train Course 3 score & Train Course 7 score\\
    \hline
    CNN+LSTM & 0.92 & 0.94 & 0.96
\end{tabular}
\end{table}

\subsection{Ablation Studies}
We explore the contributions of each component of our proposed architecture. Ablation studies were performed by removing the feature extractor, memory, and differentiator components termed remFE, remMEM, and remDIF respectively, as shown in Table \ref{table:1shotab}.  In the remFE model, the images are fed directly to the memory layer.  After trying multiple similarity metrics, the training failed.  In particular, the model diverges and predict all examples as False.  This shows the importance of the feature extractor in extracting a meaningful representation of the images for comparison.
 In the remMEM model, the feature vectors are input into the LSTM directly.  This version performed the best out of the ablation models but still does not perform as well as a more custom similarity metric.
In the remDIF model, the output from the memory layer is used as the final output.  As such, we adapted the model to use a cosine similarity metric with threshold to give a binary output.  During training, it had difficulty identifying the positive examples but still trained to convergence.  This shows the importance of the differentiator in identifying whether a stream of images matched.
\begin{table}[h]
\caption{Ablated models simulated one-shot testing results with self created metric from Section \ref{subs_perf}.}
\label{table:1shotab}
\centering
\begin{tabular}{c|c|c}
    Methods & Test course 1 score & Test course 2 score\\
    \hline
    remFE & - & -  \\
   
     remMEM & 0.02 & 0.03 \\
    
    remDIF & 0.004 & 0.017
\end{tabular}
\end{table}

\subsection{Outdoor Course} \label{outdoor}
\begin{figure*}[thpb]
      \begin{center}
      \includegraphics[width=.6\textwidth] {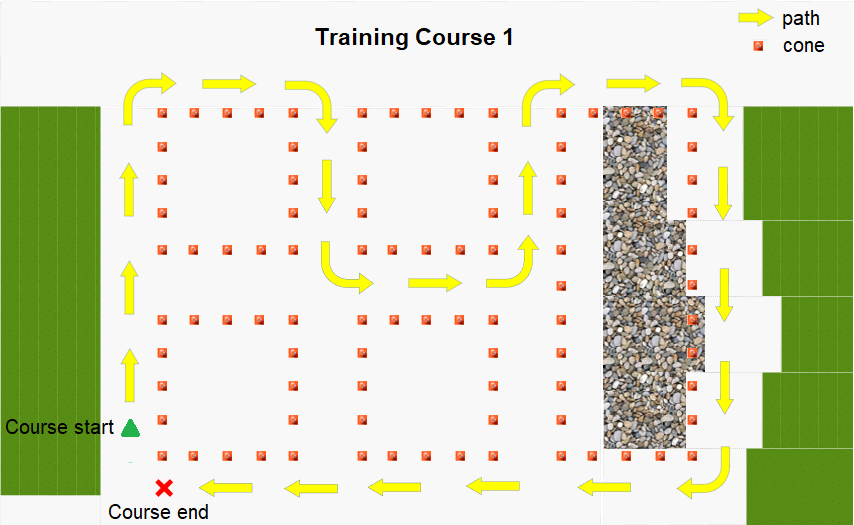}
      \caption{Example of an outdoor course}
      \label{fig:course1}
      \end{center}
    \end{figure*}

An outdoor course was set up to show the generalizability of the one-shot architecture to other driving environments.  We built a course with cones on an outdoor driveway with paths navigating the cones as shown in Figure \ref{fig:course1}.  We choose this setup because we wanted to shield the mini-AV from oncoming traffic while still processing images from road-like conditions. Because of these limitations, we could only collect data from one outdoor driveway; however, we set up the training courses and one-shot test course such that all turns but one in the one-shot test course are previously unseen. 

The result of the outdoor one-shot test course yields a passing score as in Table \ref{table:1shotoutdoor}.  This means that the one-shot test course passes in simulation, however the metric output is lower than the indoor environment.  This is likely due to less data collected for training in the outdoor environment. 
\begin{table}[!h]
\caption{Simulated one-shot testing results on outdoor course using custom metric.}
\label{table:1shotoutdoor}
\centering
\begin{tabular}{c|c}
    Method & Outdoor Test course score\\
    \hline
    CNN+LSTM & 0.56
\end{tabular}
\end{table}

\section{Conclusion}


This work has demonstrated that one-shot learning is a viable technology when few examples of a scenario are available or quick adaptation to a new environment is desired for autonomous driving applications. We developed an autonomous end-to-end DNN that can one-shot learn path navigation and tested the system online and offline in an indoor office setting and an outdoor driveway. 
While at this stage the problem is fairly simplified and the model is evaluated in a rather limited capacity, it serves as a proof of concept for the proposed layout. In a forthcoming paper, we will build on the current analysis and incorporate additional scene diversity and more sophisticated driving scenarios, as well as alternative architectures for the overall hierarchical one or few-shot model.

\begin{ack}
The authors would like to sincerely thank Nikita Jaipuria and Vidya N. Murali for fruitful discussions and for reviewing the submission draft.  We would also like to thank Nahid Pervez and Jinesh Jain for their encouragement; and Raju Nallapa for his insightful advice and continuous support.

This work was supported by Ford Motor Company.
\end{ack}


{
\small

\bibliographystyle{IEEEtranN}
\bibliography{references}
}

\newpage
\appendix
\section{Vehicle Screen Example}
\begin{figure}[h]
		\centering
		\includegraphics[width=\columnwidth]{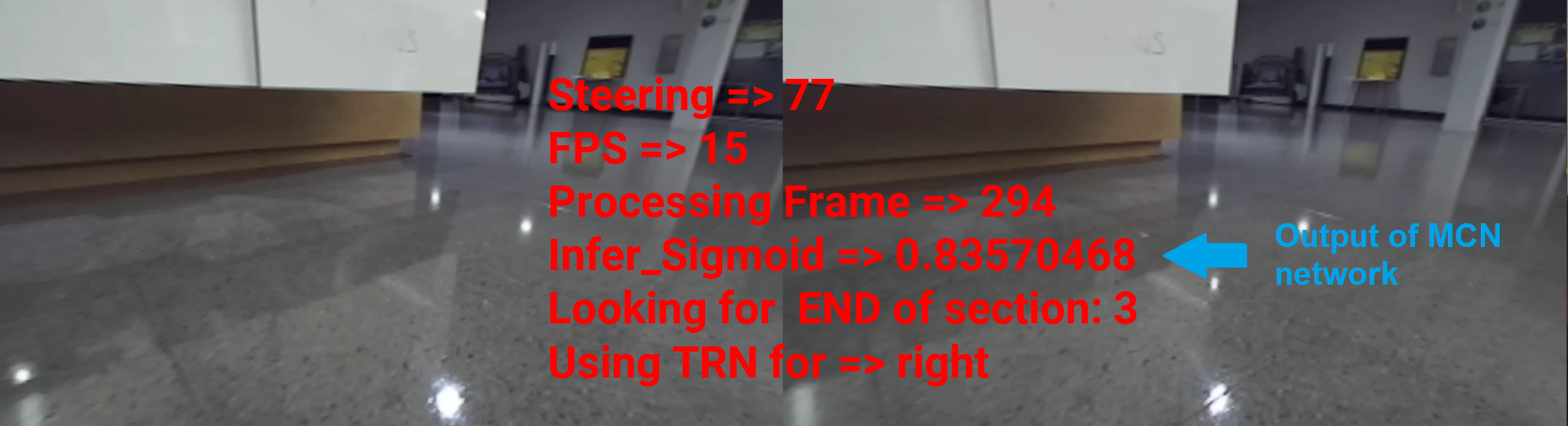}
    		\caption{Snapshot of mini-AV interface from real-time one-shot test}
		\label{fig:realrun}
	\end{figure}

A snapshot of the screen of the vehicle while driving is shown in Figure \ref{fig:realrun}. The Infer Sigmoid output is the binary output of the model and decides the path to navigate. This was captured at the end of section 3 and the high \textit{Infer\_Sigmoid} output indicates the MCN network has correctly recognized the end of section.
	
A recording of the screen of the mini-AV driving autonomously after one-shot learning is shown in the video presentation.

\section{Data Related}
\subsection{Data Pre-processing}

\begin{figure}[h]
		\centering
		\includegraphics[width= .8\columnwidth]{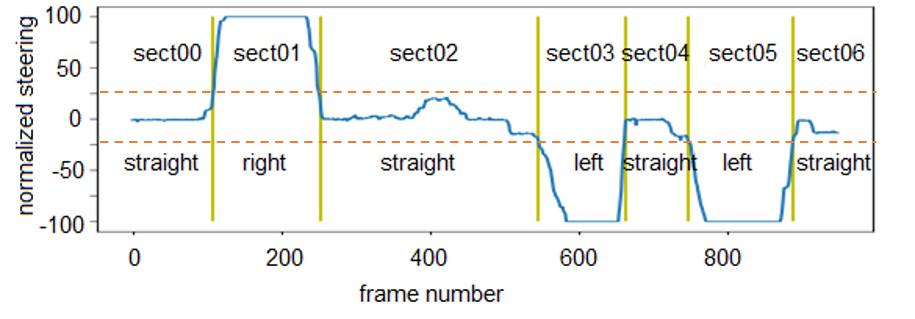}
    		\caption{Example of the section splitting algorithm; the blue curve represents the normalized steering values, the yellow lines denote the start of a new section, and the red dotted lines are guides for defining the start of a section}
		\label{fig:steerex}
	\end{figure}

There are several automated pre-processing steps performed on the raw training data. The stereo images are resized from 672x376 to 84x47 and have 3 color channels per image for a total of 6 channels.  The raw steering values are linearly mapped to normalized steering values between -100 (a full left turn) to 100 (a full right turn).  Each image frame is labeled as belonging to a ‘left’, ‘right’ or ‘straight’ section according to the following algorithm:  each frame is by default labeled ‘straight’, the turning frames are identified as a consecutive sequence of frames with steering values above 25 or below -25 and with a peak steering value of at least 90 or -90.  The frames are then split into sections based on their labels.  An example of this splitting algorithm is shown in Figure \ref{fig:steerex}.

\subsection{Data Pairing for Training}
Each training example is composed of a pair of reference and test frames, a consecutive sequence of ten frames each.  
There is a total of about 300,000 training examples. During training, 15\% is set aside as validation data. The training data has a batch size of 8, where each batch contains shuffled positive and negative examples from the same section of the same course. 

The positive and negative examples are constructed in the following way. 
 Positive examples are pairs of reference and test frames where the DNN should predict 1, i.e. both the reference and test frames are from the section end. Negative examples are pairs of reference and test frames where the DNN should predict 0, i.e. the reference frames are from the section end but the test frames are not. 

The reference frames are constructed in the same way for both positive and negative examples. The reference frames are any consecutive sequence of ten frames from the section end (the section end consists of the last fifteen frames of a section). 

The positive test examples are constructed as follows: for a particular reference frame, the test frame is any consecutive sequence of ten frames from the same section end of the same course but potentially different runs.  All valid combinations of reference and test frames are included in the training examples. 
Negative test examples are randomly sampled, and the test frame and reference frame are again from the same section of the same course but potentially different runs. The test frame is a consecutive sequence of ten frames randomly sampled from the beginning of the section up to 10 frames before the section end.  We have a 10 frame training buffer region to allow for slight differences during the manual drives.

The number of negative examples is chosen to be double the number of total positive examples because there is more diversity in the negative examples. This ratio is satisfactory in giving the model a somewhat balanced training data to identify the section end.

\subsection{Data Augmentation}
One-shot learning applications are trained on a small amount of data, so augmenting the available data to help the model generalize is particularly important. We tried various augmentation methods such as mirroring, cropping, and rotating training videos.  Each augmented course is considered a separate course from the original counterpart.  We also tested doubling the amount of near-negative training examples in each epoch, where a near-negative example is defined as a training example where the reference images are in the section end and the test images are within 15 frames of the section end.

We evaluate the effects of these augmentation methods by training 10 models for each augmented dataset and averaging their test 1 scores with the results shown in Table \ref{table:aug}.  We set the random seed for each model so that the weight initializations are the same and allow for better comparisons across methods. Unfortunately, all models trained on rotated images diverged during training. We suspect this may be due to the rotation angle being too large, and the resulting features from those rotated images being too different from the original ones.

\begin{table}[h]
\caption{Data augmentation results averaged across 10 models}
\label{table:aug}
\centering
\begin{tabular}{c|c}
     Data Aug. Methods & Average Test course 1 score \\
    \hline
    Base Data & 0.0636 \\
    
    Base, Mirrored & 0.114\\
    
    Base, Mirrored, Cropped & 0.125\\
    
    Base, Mirrored, near-negs & 0.143\\
\end{tabular}
\end{table}

These data augmentation experiments were performed before fully optimizing our model so the model scores are lower.  However, these results are averaged over 10 trainings each, so they are still meaningful and indicate that adding mirrored and near-negative data significantly increases the performance of our model. Other methods to increase training data such as additional augmentation methods or manually collecting more data may further improve our model's performance.

\section{Architecture Search}
We find the best hyperparameters and architecture styles for the manager classifier network.
    \begin{figure*}[thpb]
      \begin{center}
      \includegraphics[width=.95\textwidth, height=6cm]{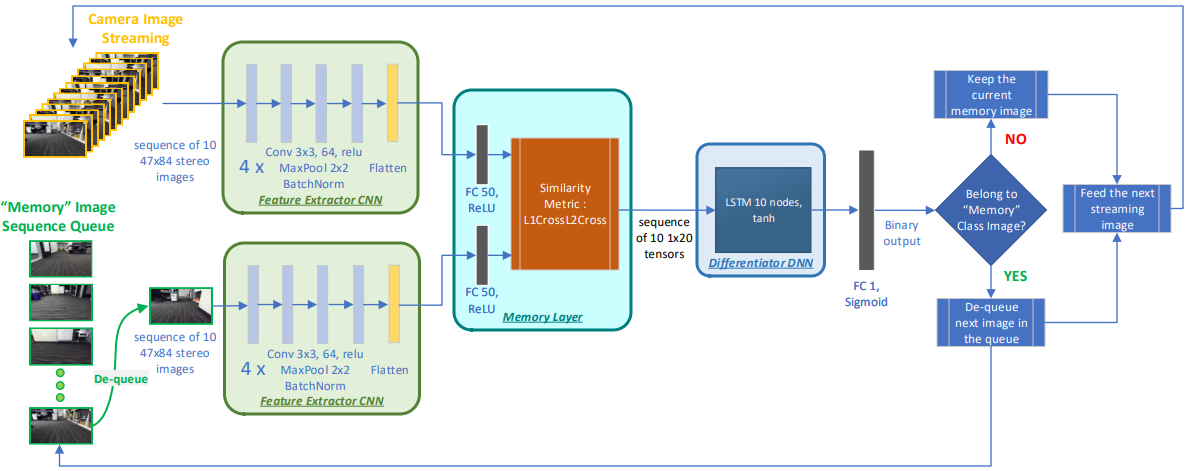}
      \caption{Overview of Manager Classifier Network and its three components: feature extractor, memory layer, and differentiator}
      \label{fig:MCN2}
      \end{center}
    \end{figure*}

\subsection{CNN Studies}
We conducted studies on modifying the CNN network (feature extractor) by changing the input image size, number of layers, filter size, and number of filters.

\subsubsection{Image Size}
We investigated the effect of image size on the output. We compared input image sizes of 47x84 and 94x168. The smaller image size produced comparable one shot test results as the larger image size, but required only 1/2 the network size and 1/3 of the training time. We concluded that the smaller image size is the best option to use with our architecture.
\begin{table}[h]
\caption{CNN study - image size}
\label{table:imagesize}
\centering
\begin{tabular}{c|c|c}
    Methods & Test course 1 score & Test course 2 score\\
    \hline
    47x84 &\textbf{0.83} & \textbf{0.96}  \\
   
    94x168 & 0.82 & 0.92 \\
    
\end{tabular}
\end{table}

\subsubsection{Number of Layers}
We experiment with CNNs composed of 4, 5, and 6 layers respectively. We find that 4 layers of CNN performs the best. For 6 layers, the validation accuracy is low and the validation and training loss are high, indicating that the model is not fitting the training or validation data well. This may be the case because in each layer, feature vector length is reduced by a max pooling layer, thus after 6 layers, the output feature vector length is ~150 compared to ~1500 after 4 layers. This reduction in feature length limits the amount of information that can be captured about the image and the distance layer performs poorly as a result.

\begin{table}[h]
\caption{CNN study - number of layers}
\label{table:nlayers}
\centering
\begin{tabular}{c|c|c}
    Methods & Test course 1 score & Test course 2 score\\
    \hline
    4 layers &\textbf{0.83} & \textbf{0.96}  \\
   
    5 layers & 0.35 & 0.86 \\
    
    6 layers & 0 & 0.01 \\
\end{tabular}
\end{table}

\subsubsection{Filter Size}
We next experiment with CNNs of varying filter size: 3x3, 5x5, or 7x7. During training and validation, they all perform well. However, the 3x3 filter size performs the best in the one shot tests.
\begin{table}[h]
\caption{CNN study - filter size}
\label{table:filtersize}
\centering
\begin{tabular}{c|c|c}
    Methods & Test course 1 score & Test course 2 score\\
    \hline
    3x3 &\textbf{0.83} & \textbf{0.96}  \\
   
    4x4 & 0.78 & 0.93 \\
    
    5x5 & 0.14 & 0.39 \\
\end{tabular}
\end{table}

\subsubsection{Number of Filters}
We next experiment with CNNs of varying number of filters of 32, 64, and 128. During training and validation, they all perform well. During the one-shot tests, CNN with 32 and 64 filters perform best in the first test while CNN with 128 filters performs best in the second test. The CNN with 64 filters performs consistently well on both tests, close to the best performance accuracy.

\begin{table}[h]
\caption{CNN study - number of filters}
\label{table:nfilters}
\centering
\begin{tabular}{c|c|c}
    Methods & Test course 1 score & Test course 2 score\\
    \hline
    32 &\textbf{0.84} & 0.94  \\
   
    64 & 0.83 & 0.96 \\
    
    128 & 0.38 & \textbf{0.97}\\
\end{tabular}
\end{table}

\subsubsection{Transfer Learning and CNN Summary}
We also considered transfer learning with a known network such as VGG16.  However, the transfer learning models, where the last 1-3 layers were finetuned in turn with our collected data did not converge during training.  We hypothesize that our collected data may be too simple for sophisticated deep networks.  It is also possible that these networks do not work well with the rest of the model architecture such as in a Siamese network or with a downstream LSTM.  Further investigation is required to pinpoint the reason.

The previous experiments indicate the best CNN design is 4 layers of 3x3 filters and with filter size of 64 and with 47x84 input image size.  

\subsection{Memory Layer Studies}
\subsubsection{Distance \& Feature Length Study}
The test and teach features outputted from the feature extractor gets compared in the memory layer.  We investigated the effects of feature length and distances on the final output. For feature length, we tried lengths of 25, 50 and 100 nodes. For each of these feature lengths, the following distances were tested with the one shot test in test course 1:
\begin{itemize}
\item Difference vector
\item Absolute difference vector
\item Square difference vector
\item Absolute and Square difference vectors concatenated
\item L1
\item L2
\item Cosine Similarity
\item L1 and L2 concatenated
\item L1 and Cosine Similarity concatenated
\item L2 and Cosine Similarity concatenated
\item L1, L2 and Cosine Similarity concatenated
\item L1 Cross Sequence
\item L2 Cross Sequence
\item Cosine Similarity Cross Sequence
\item L1 and L2 Cross Sequence concatenated
\item L1 and Cosine Similarity Cross Sequence concatenated
\item L2 and Cosine Similarity Cross Sequence concatenated
\item L1, L2 and Cosine Similarity Cross Sequence concatenated
\end{itemize}
The inputs are an image sequence, the difference between \emph{Ln} and \emph{Ln cross} is that \emph{Ln} takes that distance between each paired test and teach frames in the sequence, whereas \emph{Ln cross} takes that distance between every pair of test and teach frames.  The cross distances are more effective likely because it allows for more possibilities in matching between test and teach frames. 

\begin{table}[h]
\caption{Memory layer study - number of filters}
\label{table:nfilterslong}
\centering
\begin{tabular}{c|c|c|c}
    Distance type & Mem size 25 & Mem size 50 & Mem size 100\\
 \hline
absDiff & - &	0.04 &	0.01 \\
sqDiff&	-&	\textbf{0.84}&	0.37\\
absDiff AND sqDiff	&-	&0.39	&0.16\\
L1&	0.25	&0.38	&0.17\\
L2	&0.38	&0.27	&0.15\\
cosSim&	0.16&	0.07	&0.17\\
L1 AND L2	&0.32	&0.13	&0.07\\
L1 AND cosSim	&0.32	&0.29	&0.03\\
L2 AND cosSim	&0	&0.26	&0.00\\
L1 AND L2 AND cosSim	&0.35	&0.35	&0.06\\
L1 cross	&0.36	&0.61	&0.01\\
L2 cross	&0.63	&0.73	&0.01\\
cosSim cross	&0.31	&0	&0.01\\
L1 cross AND L2 cross	 & 0.36	&\textbf{0.83}	&0.09\\
L1 cross AND cosSim cross	&0.63	&0.12	&0.00\\
L2 cross AND cosSim cross &	0	&0.16	&0.00\\
L1 cross AND L2 cross AND cosSim cross	&0.33	&0.2	&0.00\\
L1 cross AND L2 cross AND cosSim cross	&0.33	&0.08	&0.01\\
L1 cross AND L2 cross AND cosSim cross	&0.15	& -	&0.01\\
sqDiff and L1 cross AND L2 cross &	0.3	&0.06	&0.01\\
\end{tabular}
\end{table}
The results show that memory length of 50 nodes produces the best result for the one shot test. \emph{Cosine Similarity}, a commonly used distance, does not seem to work well for this application. This is likely because cosine similarity measures the angle between two vectors but not the length of the vectors, thus two vectors can be very similar under cosine similarity but very different spatially if one vector norm is much greater than the other. \emph{L1 cross} and \emph{L2 cross} distances can work but do not yield the best results. \emph{L1 cross} and \emph{L2 cross} concatenated yielded the best results overall. This makes sense since the "cross sequence" part inserts some additional time dependency within the memory layer.

\emph{L1 cross} and \emph{L2 cross}, along with square difference vectors produced separate models that performed well on one shot test 1. However, when these 3 distances were concatenated, the resulting models performed poorly on one shot test 1. Therefore, the effects of these distances do not seem additive. \emph{L1 cross} and \emph{L2 cross} was chosen over square difference as the final metric because both \emph{L1 cross} and \emph{L2 cross} by themselves also performed well, so we postulate that these two together is a more stable distance.  Other distances may be investigated to improve our results as future work.

\section{Outdoor Driving Environment}

We have shown that the mini-AV can learn a path in one-shot in an indoor environment.  We also collected data from an outdoor course to show the generalizability of this one-shot architecture.

\subsection{Course}
\begin{figure}
      \begin{center}
      \includegraphics[width=\columnwidth]{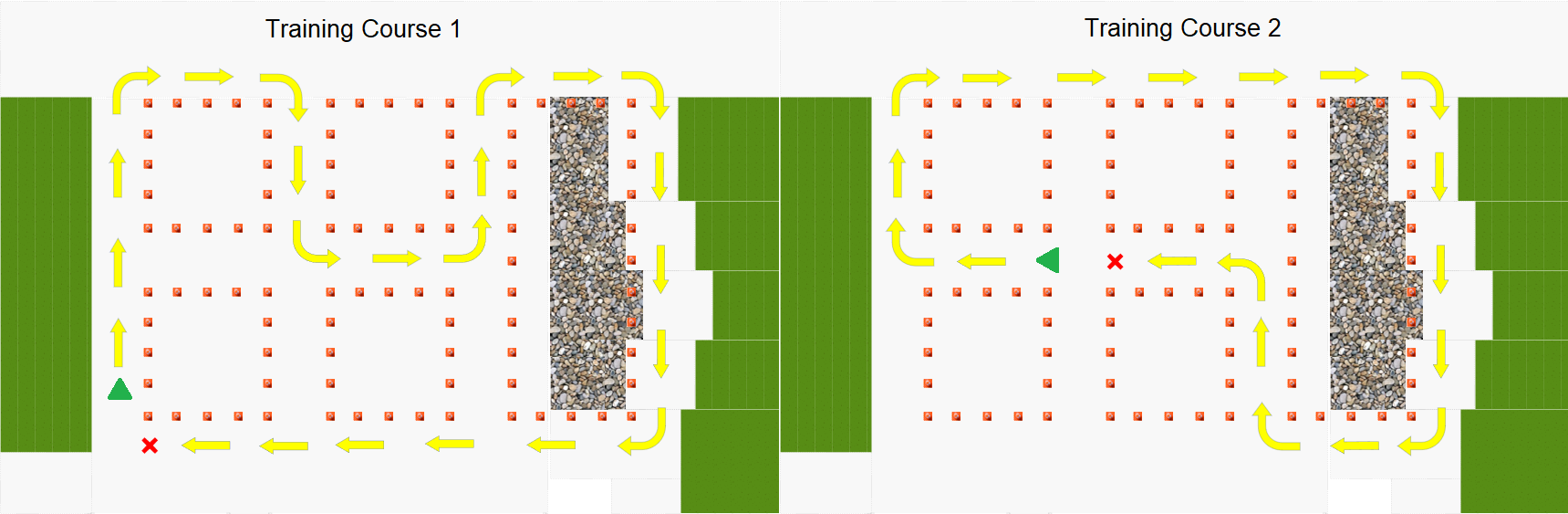}
      \caption{training courses}
      \label{fig:trainingout}
      \end{center}
    \end{figure}

\begin{figure}
      \begin{center}
      \includegraphics[width=.5\columnwidth]{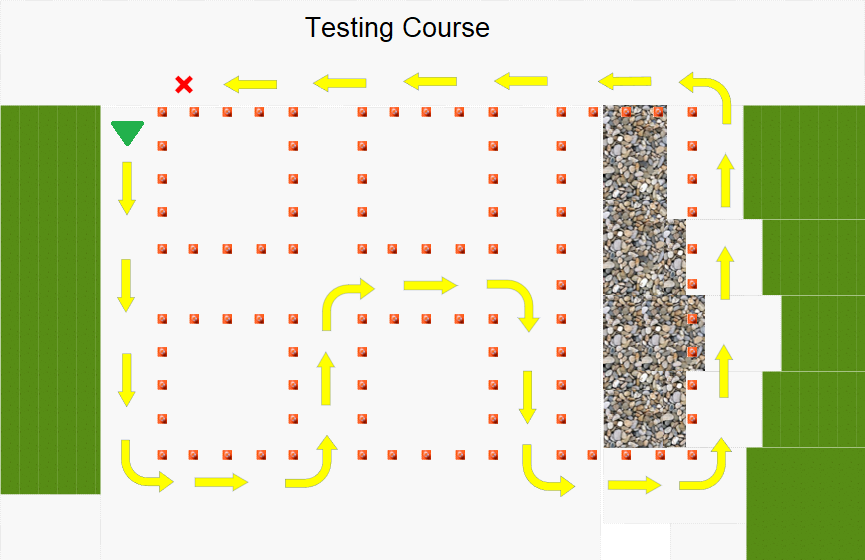}
      \caption{one-shot test course}
      \label{fig:testout}
      \end{center}
    \end{figure}

We built a course with cones on an outdoor driveway.  We choose this location because we wanted to make sure the mini-AV is shielded from oncoming traffic while still processing images from road-like conditions.
Because of these limitations, we could only collect data from one outdoor driveway, however we set up the training courses (Figure \ref{fig:trainingout}) and one-shot test course (Figure \ref{fig:testout}) such that all but one turn in the one-shot test course is previously unseen.  The training courses are again collected with 3 runs per course.  The 2 courses contributed a total of 5837 training batches.

\section{Performance Metric}
We developed a novel performance metric to evaluate the One-Shot Learning model with regards to its overall accuracy for predictions during a One-Shot Test as well as the overall success of the RC car steering system. For offline testing, a simulated One-Shot test was used and evaluated.

The evaluation One-Shot Learning test is performed using data from a course that the model has never seen before (for indoors, a course from Building A). Two files are required for testing. The first file serves as the "One-Shot" experience, and the images from this file are used as the model's "memory." The second file represents a second pass through the course, and this time the model uses its "memory" to compare the images from the first file to the incoming images in the second file.

For each frame in the second file, the model predicts whether the frame is a match for the corresponding end-of-section memory frames.

    \begin{figure*}[thpb]
      \begin{center}
      \includegraphics[width=.8\textwidth]{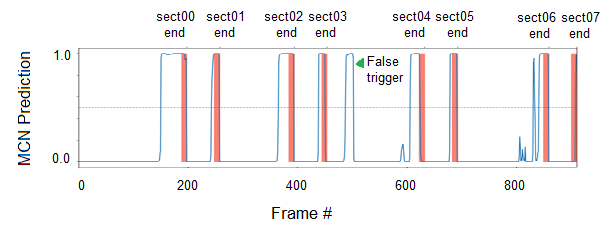}
      \caption{Example of a failed test output}
      \label{fig:badex}
      \end{center}
    \end{figure*}
An overall performance metric is needed because simple measures such as accuracy do not tell the full story.  For example in Figure \ref{fig:badex}, the overall accuracy in the section bodies is $97\%$, but it is obvious that in the body of sect04, there is a False trigger.  In our setup, this would cause the mini-AV to prematurely make a turn and fail the test path.  This catastrophic failure is not reflected in the accuracy thus we design the performance metric to output a low score for every failure in the section body and/or section end.  The following sections covers the performance metric components in detail.

\subsection{Metric Breakdown}

The performance metric considers an evaluation run as a whole, as well as on a section-by-section basis. It is composed of several sub-metrics: Section Body Accuracy, Section End Accuracy, Failure Boolean, and Trigger Success Boolean.  These metrics can be looked at cumulatively or individually for specific debugging of a particular section within the run. 

\subsection{Section Body Accuracy}
The section body accuracy is a measure that reflects the true negative rate or specificity of the section body. Since the model is expected to predict 0 during the section body, any errors here are false positives.

We consider each frame in the section body that has a prediction less than 0.5 to be counted as correct predictions. For each of these correct predictions, we calculate the "distance", or number of frames between this frame and the termination of the section body, and we add this distance to an overall summation. This overall summation is then divided by the cumulative sum of the distances between each frame and the termination of the section body for every single frame in the section body. The distance measurements from frames with erroneous predictions are not included in the numerator sum, and therefore help create an overall "accuracy" measure.

Since the model is expected to predict "1" during each section end, we consider false positives that are closer to be beginning of the section body to be worse errors than those nearer to the section end. Therefore, the distance between each frame and the last frame of the section body was chosen as a way to weight the relevance of the predictions. This causes frames with erroneous predictions that are farther away from the section end to detract more heavily from the accuracy percentage.

\begin{equation} \label{eq:3}
        \textrm{section body accuracy} = \frac{\sum_{i = start frame_j}^{n_{j}}{(n_{j}-i)\cdot \mathbbm{1}[p_i < 0.5]}}{\sum_{i = start frame_j}^{n_{j}}{(n_{j}-i)}}
\end{equation}

\subsection{Section End Accuracy}

The section end accuracy is used to assess the performance of the model during each section end. Since the model is expected to predict '1' for each frame in the section end, this metric reflects the true positive rate or sensitivity of the section end.

We consider each frame in the section end with a prediction greater than 0.5 to be a correct prediction. We count the number of correct predictions within the section end and then divide by the total number of frames in the section end to produce an accuracy measure as seen in Figure \ref{endAccuracy2}.

\begin{equation} \label{eq:4}
        \textrm{section end accuracy} = \frac{\sum_{i \in \{section end\}} \mathbbm{1}[p_i> 0.5]}{\textrm{total number of frames in section end}}\label{endAccuracy2}
\end{equation}

\subsection{Failure Boolean}

The Failure Boolean is a binary measure of system failure, based on the model's predictions throughout each section body. 

Failure is defined as a false positive prediction that extends for a given length of frames. In these experiments, the threshold prediction value of 0.5 is used, as well as the threshold length value of 5 frames. The threshold length value is used everywhere in the test run except for checking false positive prediction stretches that extend directly into the section ends. For these cases, a threshold length value of 10 frames is used.

The algorithm tracks false positive predictions throughout each section body and flags and records sequential false positive predictions. This information is used to determine if the run meets the failure criteria of containing more than 5 subsequent frames with false positive predictions.

\subsection{Trigger Success Boolean}

The Trigger Success Boolean is a binary measure of if the system is expected to "trigger" and switch steering networks correctly, based on its predictions during each section end.

To complement the previous definition of failure in the section body, success in the section end is defined as a prediction above a given threshold for a given length of frames. Here, we use the threshold value of 0.5 and the threshold length value of 5 frames.

To include the possibility that the systems triggers a few frames earlier than the section end, which has no overall impact in the system performance, a test buffer region is set before the section end where a trigger is valid in as well. This buffer region is equal to the length of the section end (in these experiments, it was 15 frames).

\subsection{Combined Metric}
As mentioned in the main body of this paper, the combined metric applies the geometric mean several times to the section body accuracy and section end accuracy. We also considered variations such as taking the arithmetic mean several times or in combination with the geometric mean, and using the specificity directly as the section body accuracy.  By observing the results and cross-checking with the simulated output graphs, we found the proposed metric captured the quality of the predictions the best.

\end{document}